\documentclass[10pt,twocolumn,letterpaper]{article}
\usepackage{cvpr}             

\usepackage{graphicx}
\usepackage{amsmath}
\usepackage{amssymb}
\usepackage{booktabs}
\usepackage{multirow}
\usepackage{adjustbox}

\usepackage[pagebackref,breaklinks,colorlinks]{hyperref}

\usepackage[capitalize]{cleveref}
\crefname{section}{Sec.}{Secs.}
\Crefname{section}{Section}{Sections}
\Crefname{table}{Table}{Tables}
\crefname{table}{Tab.}{Tabs.}

\begin{document}
\title{PTA-Det: Point Transformer Associating Point cloud and Image\\ for 3D Object Detection}
\author{Rui Wan\textsuperscript{\rm 1}, Tianyun Zhao\textsuperscript{\rm 1\thanks{Corresponding author}*}, Wei Zhao\textsuperscript{\rm 2}\\
\textsuperscript{\rm 1}School of Automation, Northwestern Polytechnical University \\
\textsuperscript{\rm 2}Institute of Photonics \& Photon-Technology, Northwest University \\
{\tt\small zhaoty@nwpu.edu.cn}
}
\maketitle 
\begin{abstract}
In autonomous driving, 3D object detection based on multi-modal data has become an indispensable approach when facing complex environments around the vehicle. During multi-modal detection, LiDAR and camera are simultaneously applied for capturing and modeling. However, due to the intrinsic discrepancies between the LiDAR point and camera image, the fusion of the data for object detection encounters a series of problems. Most multi-modal detection methods perform even worse than LiDAR-only methods. In this investigation, we propose a method named PTA-Det to improve the performance of multi-modal detection. Accompanied by PTA-Det, a Pseudo Point Cloud Generation Network is proposed, which can convert image information including texture and semantic features by pseudo points. Thereafter, through a transformer-based Point Fusion Transition (PFT) module, the features of LiDAR points and pseudo points from image can be deeply fused under a unified point-based representation. The combination of these modules can conquer the major obstacle in feature fusion across modalities and realizes a complementary and discriminative representation for proposal generation. Extensive experiments on the KITTI dataset show the PTA-Det achieves a competitive result and support its effectiveness.
\end{abstract}
	\begin{figure}
		\centering
		\includegraphics[width=0.95\columnwidth ,height=5.7cm]{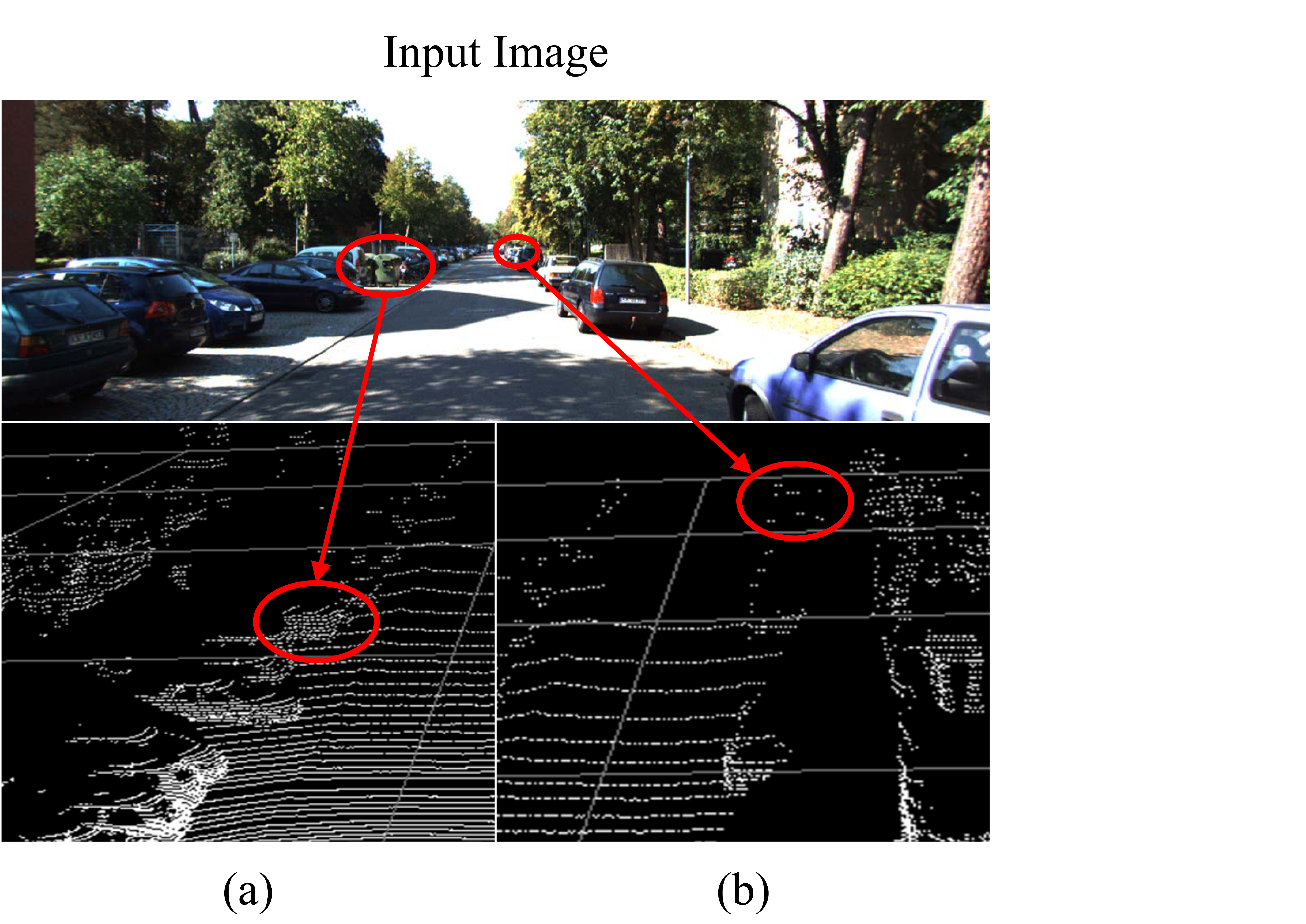}
		\vspace{-3mm}
		\caption{(a) The background obstacle has a similar 3D structure to a car. It is difficult to distinguish them in the point cloud. (b) The points of the distant car are too sparse to determine the bounding box.}
		\vspace{-5mm}
		\label{fig:Visio-illustration}
	\end{figure}
\section{Introduction}
3D object detection is a fundamental perception task for autonomous driving, which usually takes point clouds or images as input. It aims to estimate the 3D bounding boxes of objects and recognize their categories. Due to the success of convolution neural network (CNN) in 2D object detection, image-based 3D object detection has emerged to obtain spatial clues about the objects. Although the image elaborates the front-view projection of objects in a 3D scene, accurate depth measurement is still required in order to localize 3D objects. In the last decade, 3D sensors such as LiDAR have developed rapidly. Relying on the devices, researchers can obtain point clouds that reflect the relative positions of the sensor and the obstacle for 3D object detection. Early works relying on LiDAR points (e.g., PointRCNN \cite{DBLP:conf/cvpr/ShiWL19} and VoxelNet \cite{DBLP:journals/corr/abs-1711-06396}) have achieved superior results over image-based detection methods. However, they suffer from the poor semantic information of point clouds as shown in Figure \ref{fig:Visio-illustration}(a). In addition, the objects in Figure \ref{fig:Visio-illustration}(b) are difficult to detect using LiDAR-only methods because the distant point cloud is extremely sparse. In contrast, an image is an array of pixels and can provide continuous textural structure, which is in favor of distinguishing false-positive instances. Intuitively, it is essential to design a multi-modal 3D object detection method to exploit both the geometry clues in the point cloud and the textural clues in the image.
\par As can be seen from the KITTI \cite{DBLP:conf/cvpr/GeigerLU12,DBLP:journals/ijrr/GeigerLSU13} leaderboard, there is still a gap between the mean Average Precision (mAP) of multi-modal detection methods and LiDAR-only methods. \textbf{The first performance bottleneck encountered by multi-modal methods is the capability to extract intra-modal features.} According to the related works \cite{DBLP:journals/corr/ChenMWLX16,DBLP:conf/iros/KuMLHW18,DBLP:conf/cvpr/QiLWSG18,DBLP:conf/cvpr/XuAJ18,DBLP:conf/icra/SindagiZT19,DBLP:conf/cvpr/VoraLHB20,DBLP:journals/corr/abs-1903-01864,DBLP:journals/corr/abs-1911-06084,DBLP:journals/corr/abs-2007-08856,DBLP:conf/cvpr/QiCLG20,DBLP:journals/corr/abs-2112-11088,DBLP:conf/cvpr/Wang0ZY21,CHEN202223}, normally, the features of point cloud are extracted via PointNet++ \cite{DBLP:conf/nips/QiYSG17}/3D sparse convolution \cite{DBLP:journals/sensors/YanML18}, while image features are extracted through 2D convolution \cite{NIPS2012_c399862d}. But the useful long-distance dependencies in each modality are difficult to capture due to the local receptive fields of these building blocks. To balance speed and accuracy, multi-modal methods usually reduce the number of point clouds and image size of the input. The approach will cause serious long-distance information loss and reduce detection accuracy.
\par\textbf{The second performance bottleneck of multi-modal methods is restricted by the fusion method of inter-modal features.} Frustum-PointNet \cite{DBLP:conf/cvpr/QiLWSG18} generates 2D proposals for images and lifts them into frustums. It predicts the bounding box from points in the extruded frustum. As proposed in the investigation \cite{DBLP:conf/cvpr/WangTF20}, using multiple data separately does not take their complementarity into account. AVOD \cite{DBLP:conf/iros/KuMLHW18} projects point clouds into Bird’s eye view (BEV) and aggregates BEV features and image features in anchors to generate proposals. Owing to the quantized errors during projection, the result is affected by an inaccurate alignment between the two features. PointPainting \cite{DBLP:conf/cvpr/VoraLHB20} performed image semantic segmentation and appends the semantic score of the projected position on the image to the point cloud features. The enhanced features are sent to point-based method for proposal generation. Although the mAP is relatively improved, the simple feature concatenation is not enough to fuse image and point cloud features. In summary, the main reason limiting their accuracy is the problematic fusion of multi-modal data. 
\par To conquer the first performance bottleneck, we strive to extract both point cloud features and image features. To extract point cloud features, the transformer \cite{DBLP:journals/corr/VaswaniSPUJGKP17} proposed in machine translation is used to construct the feature extraction module. Recent studies \cite{DBLP:journals/corr/abs-2012-09164,DBLP:journals/cvm/GuoCLMMH21} have shown that transformer is capable of point cloud classification and segmentation. Compared with CNNs, transformer is developed based on a self-attention mechanism and can capture long-distance dependencies. Therefore, relying on transformer, Point Transition Down (PTD) and Point Transition Up (PTU) modules are designed to extract point cloud features. 
\par In contrast, the extraction of image features in multi-modal detection is still an open problem. Inspired by the application of pseudo point cloud (PPC) in 3D object detection by MDPPC \cite{DBLP:conf/iccvw/WengK19}, a Pseudo Point Cloud Generation network that converts image pixels into PPCs is developed, and then PPCs are used to represent the image features. Attributed to the unified form of PPC and raw point cloud, both image features and point cloud features can be learned in the form of points. 
\par To address the second performance bottleneck, a two-stream feature extraction network, based solely on transformer, is developed to solve the fusion of inter-modal features. Specifically, the two-stream structure consists of a point cloud branch and a PPC branch. The two branches independently learn high-level point cloud features and PPC features. In particular, the image features of the object keypoints are used as the initial PPC features, and these features are further analyzed and encoded in the PPC branch. Benefiting from the unified feature representation, a Point Fusion Transition (PFT) module is designed, which accurately fuses the two features at the point level to highlight the key cues across the two modalities. 
\par In general, we present a multi-modal 3D object detection method PTA-Det, constructed on the basis of point-level feature learning and fusion. Accompanied by a series of modules, the mAP of multi-modal 3D object detection can be improved with better robustness and accuracy. 
Our main contributions are summarized as follows:
\begin{itemize}
\item The PPC generated by a Pseudo Point Cloud Generation network, a point-based representation of image feature, has been leveraged for multi-modal 3D object detection.
\item A two-stream feature extraction network entirely relying on transformer has been developed, to learn intra-modal features and inter-modal features at the point level. 
\item Competitive results on KITTI dataset have been achieved. Results demonstrate that our model is compatible with most LiDAR-only detectors and can be easily upgraded to a multi-modal detector.
\end{itemize} 
	\begin{figure*}
		\centering
	\includegraphics[width=0.98\linewidth, height=6.5cm]{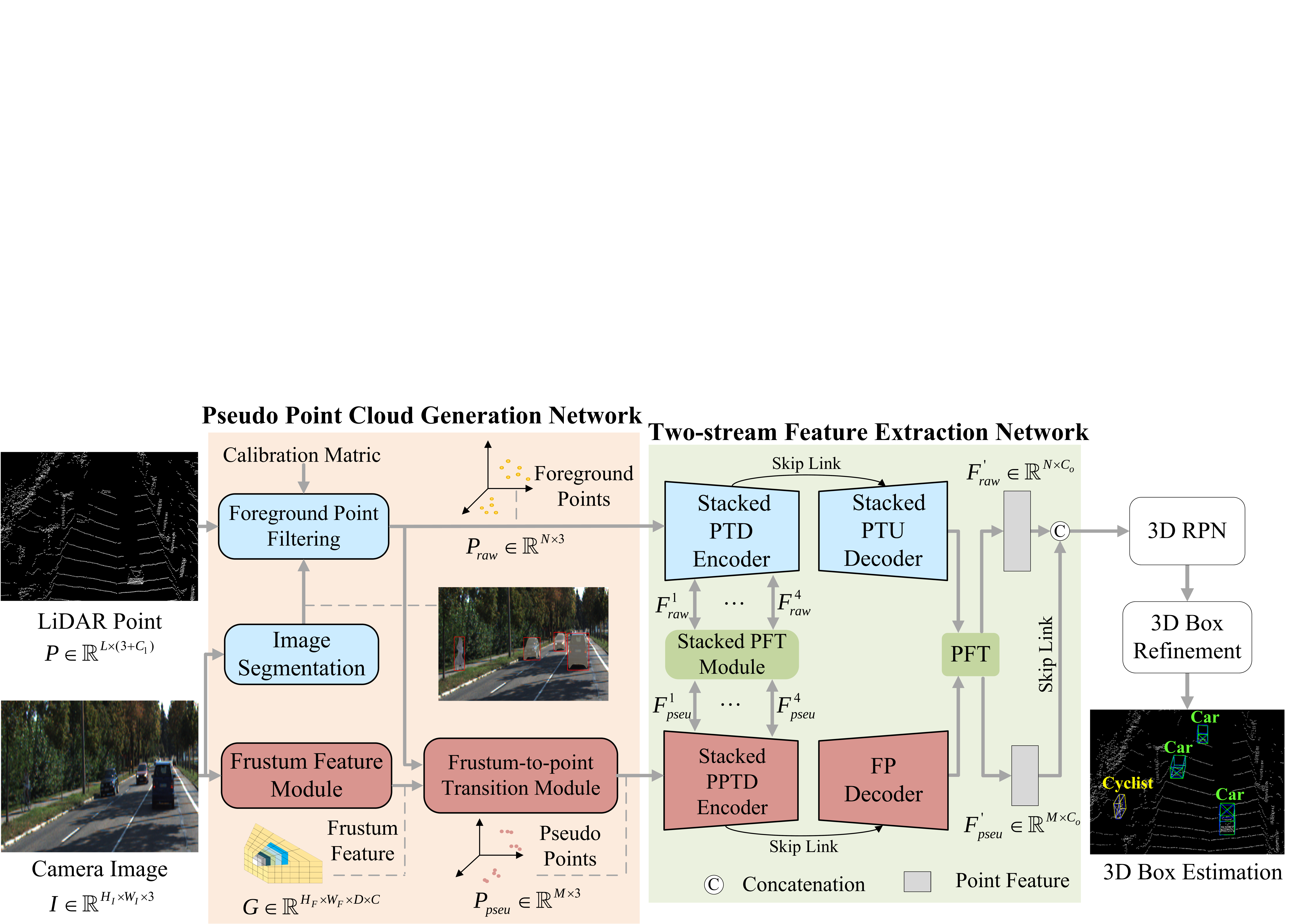}
		\vspace{-2mm}
		\caption{Illustration of the architecture of PTA-Det. It is composed of four parts: (1) Pseudo Point Cloud Generation Network for generating the coordinates and features of PPCs, (2) Two-stream Feature Extraction Network for learning point-based intra-modal and inter-modal features, (3) 3D RPN for proposal generation, (4) 3D Box Refinement for rectifying proposals. }
		\label{fig:backbone}
		\vspace{-3mm}
	\end{figure*}	
\section{Related Work}
\label{sec:RW}
\noindent\textbf{PPC-based 3D Object Detection.} From the investigation of Reading et al \cite{DBLP:conf/cvpr/ReadingHCW21}, image-based detection methods show unsatisfactory results owing to lacking direct depth measurement. However, Wang et al \cite{DBLP:conf/cvpr/WangCGHCW19} argued that image-based methods are mainly affected by the representation of the data rather than its quality. They converted the depth image to the PPCs and applied a point-based method to detect objects. Pseudo-LiDAR \cite{DBLP:conf/iccvw/WengK19} performed depth estimation and proposal prediction on image. For each proposal, a point cloud frustum is extracted from the PPCs obtained from the depth image transformation. Nevertheless, both of these methods ignore the depth error introduced into the PPC, which further affects their detection accuracy. To solve the problem, Pseudo-LiDAR++ \cite{DBLP:journals/corr/abs-1906-06310} utilized extremely sparse LiDAR points to correct their nearby PPCs to achieve accurate depth prediction. Plug-and-play \cite{DBLP:conf/icra/WangWLTCS19} proposed a PnP module that integrated sparse depth values into an intermediate feature layer to correct depth prediction. In addition to depth correction, End-to-End Pseudo-LiDAR \cite{DBLP:journals/corr/abs-2004-03080} jointly trained depth prediction and object detection for accurate proposals. Hence, in our proposed Pseudo Point Cloud Generation network, we not only dynamically generate PPCs, but also apply corresponding features for the subsequent detection pipeline.\\
\textbf{Multi-modal based 3D Object Detection.} According to the different fusion strategies, the existing multimodal detection methods can be divided into three categories: result-level, proposal-level and point-level methods. In the result-level methods \cite{DBLP:journals/corr/ChenMWLX16,DBLP:conf/cvpr/QiLWSG18,DBLP:conf/cvpr/QiLWSG18,DBLP:journals/corr/abs-1911-06084,CHEN202223}, it is common to utilize the feature of one modality to generate the proposal, and utilize the feature of the other modality in the proposal to generate the bounding box. These methods have high recall even when the object is far or occluded, but their accuracy is limited by ignoring the complementarity between different data. The proposal-level methods \cite{DBLP:conf/iros/KuMLHW18,DBLP:conf/icra/SindagiZT19,DBLP:conf/cvpr/Wang0ZY21,DBLP:journals/corr/abs-2012-12397} take the encoded features of image and point cloud as inputs, and fuse the two features in anchors to generate proposals. These methods benefit from multi-modal data and can generate high-quality proposals. However, their performance is affected by irrelevant information mixed in the anchors and inaccurate feature alignment. The point-level methods \cite{DBLP:conf/cvpr/VoraLHB20,DBLP:journals/corr/abs-2007-08856,DBLP:conf/cvpr/QiCLG20,DBLP:journals/corr/abs-2112-11088,DBLP:journals/corr/abs-2204-00325} have shown promising results. ImVoteNet \cite{DBLP:conf/cvpr/QiCLG20} fused 2D votes of image and point cloud features in a point-wise concatenation manner, but the approach is insufficient to fuse the two features. To address the drawback, EPNet \cite{DBLP:journals/corr/abs-2007-08856} proposed the LI-Fusion module that adaptively fuses point cloud features with image features according to the importance of the image feature channel. EPNet++ \cite{DBLP:journals/corr/abs-2112-11088} proposed the CB-Fusion module that added the fusion direction from the point domain to the image domain. It showed that bi-direction interaction approach leads to a more comprehensive and discriminative feature representation. Recently, several transformer-based multi-modal detection methods \cite{DBLP:journals/corr/abs-2204-00325,DBLP:conf/cvpr/BaiHZHCFT22} have been proposed. Dosovitskiy et al \cite{DBLP:conf/iclr/DosovitskiyB0WZ21} demonstrated that transformer have comparable expressive power to CNN. Therefore, we choose transformer to build the backbone network and develop a PFT module for bi-direction information interaction.
\section{Method}
\label{sec:Method}
In this research, we present a multi-modal 3D object detection method named PTA-Det. As shown in Figure \ref{fig:backbone}, the PTA-Det mainly consists of a Pseudo Point Cloud Generation Network, a Two-stream Feature Extraction Network, a 3D Region Proposal Network (RPN), and a 3D Box Refinement Network. The Pseudo Point Cloud Generation Network comprises a Frustum Feature module and a Frustum-to-point Transition module. The Two-stream Feature Extraction Network consists of a point cloud branch and a PPC branch. The former contains a stacked PTD encoder and a stacked PTU decoder, and the latter contains a stacked pseudo PTD (PPTD) encoder and a stacked FP decoder. Meanwhile, the stacked PFT module is used to connect the two branches at multiple levels. In the following, the four modules utilized in the investigation are elucidated in sequence. 
	\begin{figure*}
		\centering
	\includegraphics[width=0.82\linewidth, height=5.5cm]{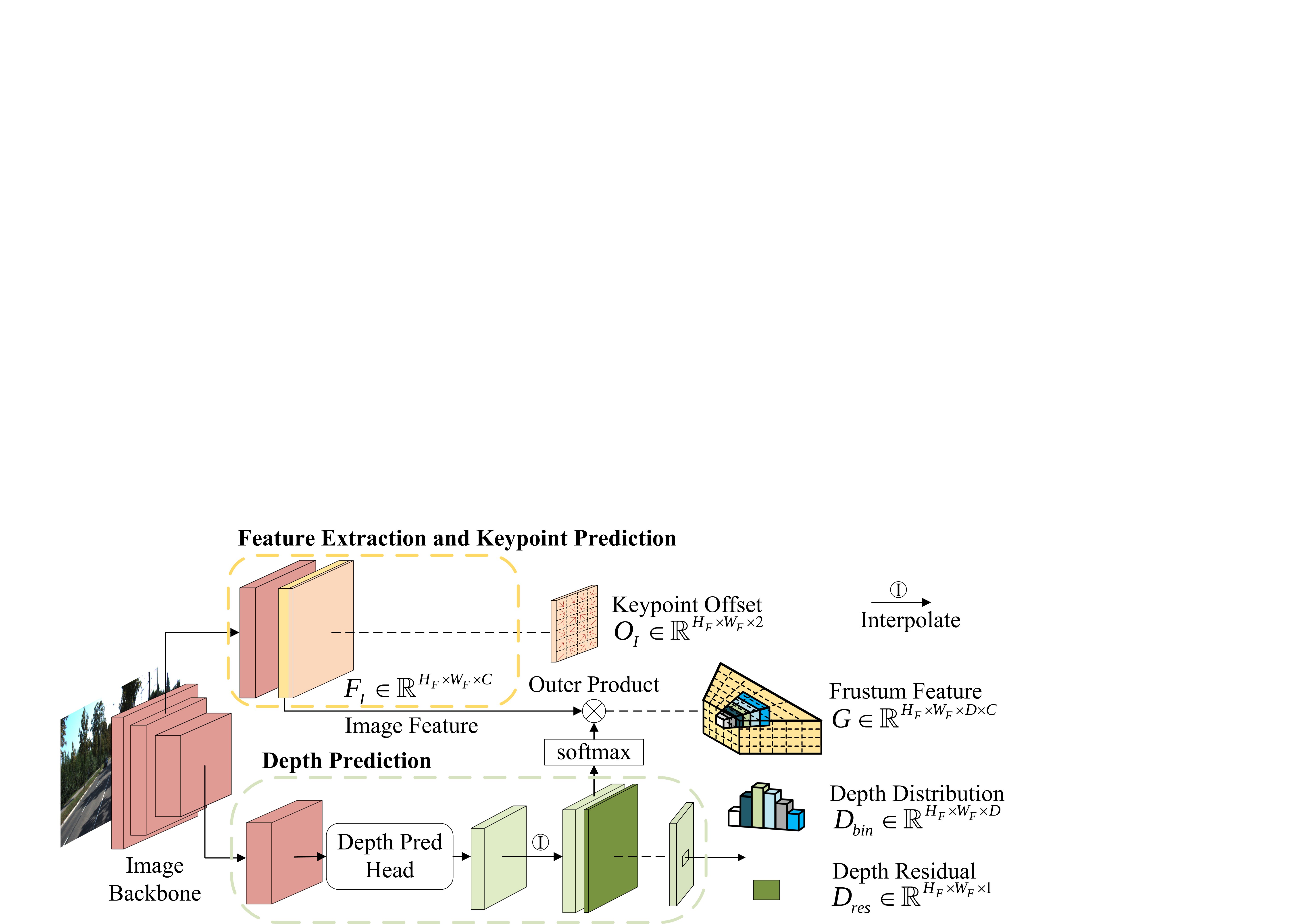}
		\vspace{-2mm}
		\caption{Illustration of Frustum feature network.}
		\label{fig:FTN}
		\vspace{-3mm}
	\end{figure*}	
\subsection{Pseudo Point Cloud Generation Network}
\label{sec:3.1}
In this network, image is transformed into PPCs which are further utilized to represent image features. During processing, the image depth is predicted in a semi-supervision manner. LiDAR points are projected onto the image to obtain sparse depth labels, which are used to supervise depth prediction. With the help of the foreground mask from Mask-RCNN \cite{DBLP:conf/iccv/HeGDG17} and the predicted depth image, the foreground pixels can be converted into PPCs. At the same time, adhering to CaDDN \cite{DBLP:conf/cvpr/ReadingHCW21}, the Frustum Feature Network is used to construct the frustum feature. Then, the PPC features are obtained by interpolating the frustum feature in the Frustum-to-point Transition module.\\
\textbf{Frustum Feature Module.} In order to make full use of the image information, a Frustum Feature module is constructed to generate frustum feature. In Figure \ref{fig:FTN}, extracting image features and predicting image depth are two fundamental steps. Similar to CaDDN \cite{DBLP:conf/cvpr/ReadingHCW21}, ResNet-101 \cite{DBLP:conf/cvpr/HeZRS16} is utilized as the backbone to process images and the output of its Block1 is used to collect image features $F_I\in\mathbb{R}^{H_F\times{W_F}\times{C}}$, where $H_F$,$W_F$ are the height and width of the image feature, and $C$ is the number of feature channels.
\par On the other hand, a depth prediction head is applied to the output of the image backbone to predict image depth. The depth prediction is viewed as a bin-based classification problem and the depth range is discretized into D bins by the discretization strategy LID \cite{DBLP:journals/corr/abs-2005-13423}. Then the depth distribution $D_{bin}\in\mathbb{R}^{H_F\times{W_F}\times{D}}$ and depth residual $D_{res}\in\mathbb{R}^{H_F\times{W_F}\times{D}}$ can be obtained.
\par Early depth estimators \cite{DBLP:journals/corr/abs-2005-13423,DBLP:conf/cvpr/ReadingHCW21,DBLP:journals/corr/abs-1907-10326} computed the loss over the entire image including a large number of background pixels. These methods place over-emphasis on background regions in depth prediction. According to Qian et al \cite{DBLP:journals/corr/abs-2004-03080}, background pixels can occupy about 90\% of all pixels in KITTI dataset. Therefore, instead of calculating the loss of all image pixels, the off-the-shelf image segmentation network Mask-RCNN \cite{DBLP:conf/iccv/HeGDG17} is employed to select N foreground points from LiDAR points by distinguishing their 2D projection positions. The N points are re-projected onto the image to acquire sparse depth label for calculating the depth loss of foreground pixels. In addition, the foreground loss will be given more weight to balance the contributions of foreground and background pixels. 
\par With the image feature and image depth, the frustum feature $F_T\in\mathbb{R}^{H_F\times{W_F}\times{D}\times{C}}$ can be constructed as follows
\begin{equation}\label{eq1}
    F_T=SM(D_{bin})\otimes{F_I}
\end{equation}
where $\otimes$ is the outer product and $SM$ represents the $SoftMax$ function. Eq. \ref{eq1} states that at each image pixel, the image features are weighted by the depth distribution values along the depth axis. CNN is known to extract image features in convolutional kernels, where object pixels may be surrounded by the pixels of the background or other objects. In contrast, the frustum feature network lifts image features onto depth bins along the depth axis, which enables the model to discriminate misaligned features in 3D space.\\
\textbf{Frustum-to-point Transition Module.} The subnetwork aims to extract the features of PPC from frustum feature. There are two issues to be addressed regarding the choice of PPC. First, due to the presence of depth errors, the PPCs converted from image may not be consistent with the distribution of the object in space. Second, the number of PPC is proportional to the image resolution. Nevertheless, only in the area where the point cloud is relatively sparse, PPC can play an important role by compensating for the missing object information. 
	\begin{figure}
		\centering
	\includegraphics[width=0.88\columnwidth]{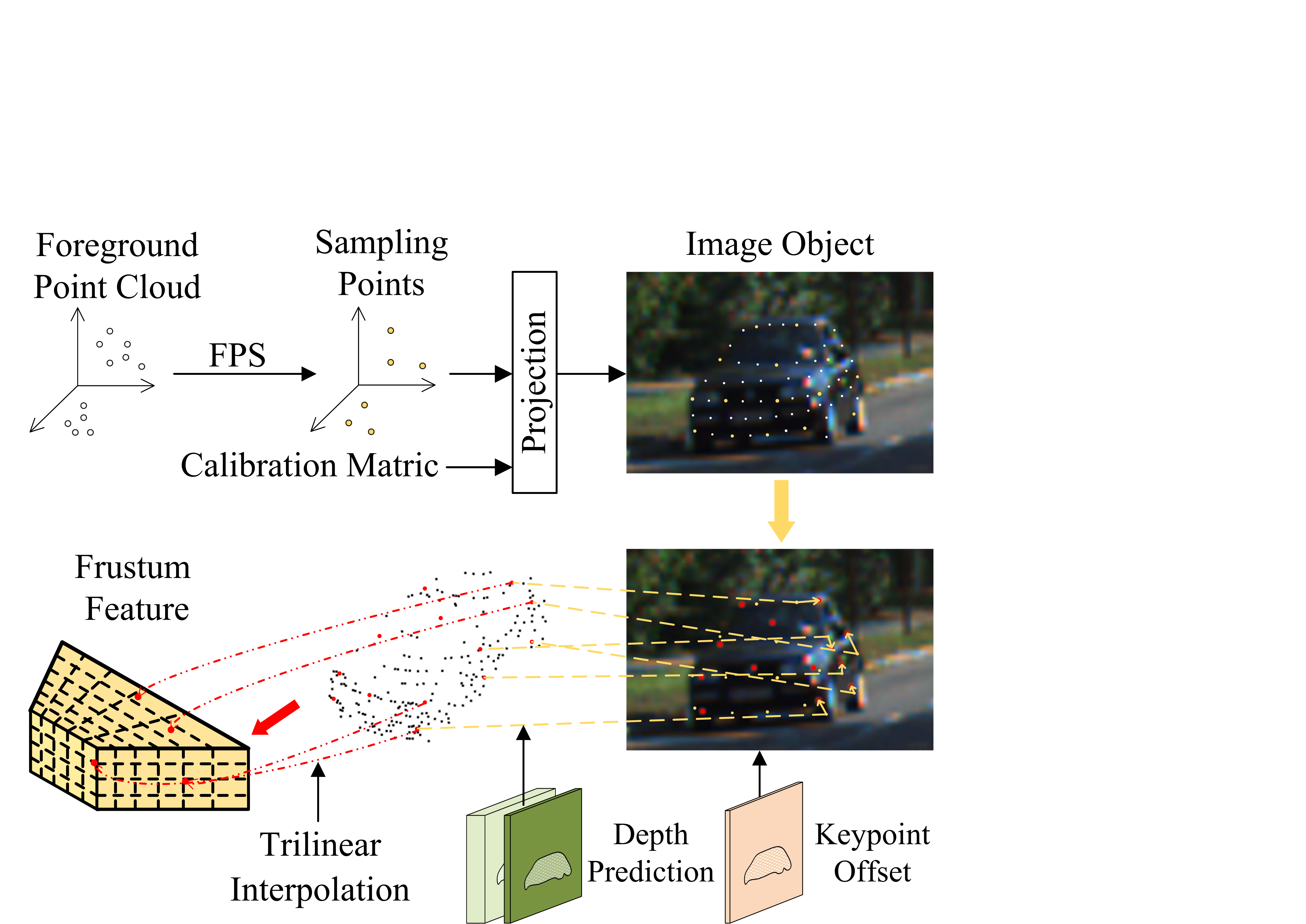}
		\vspace{-3mm}
		\caption{Illustration of Frustum-to-point Transition module. The white point is the foreground point, yellow is sampled foreground point, and red is the keypoint of the object.}
		\vspace{-5mm}
		\label{fig:FtP}
	\end{figure}
\par For the first issue, we apply the farthest point sampling (FPS) algorithm to select M points from the N foreground points as the initial PPCs in Figure \ref{fig:FtP}. Foreground points are used because they have more accurate depth values near the projected positions in the image. Accordingly, the projected coordinates $\{c_i=(u_i,v_i)\vert{i=1,\dots,M}\}$  of M foreground points can be obtained via calibration matrix.
\par As for the second issue, the object keypoints that focus on more representative object information are introduced as the final PPCs. Keypoints are defined as locations that reflect the local geometry of an object, such as points on mirrors and wheels. To determine the locations of keypoints in 3D space, inspired by Deformable Convolutional Networks \cite{DBLP:conf/iccv/DaiQXLZHW17}, a 2D keypoint offset is predicted which represents the offset of each pixel on the image to its nearest keypoint. For the M projected coordinates, M keypoint offsets are acquired as follows
\begin{equation}
    O_I(c_i)=(\Delta{u_i},\Delta{v_i})=\sum\nolimits_{q}G(c_i,q)O_I(q)
\end{equation}
where $q$ enumerates the nearby integral locations of $c_i$ on the image and $G(\cdot,\cdot)$ is the bilinear interpolation kernel. Keypoint offset $O_I\in\mathbb{R}^{H_F\times{W_F}\times{2}}$ is predicted when generating image features illustrated in Figure \ref{fig:FTN}.
\par Then, the locations of the 2D keypoints can be obtained as $\{c_i^{'}=(u_i+\Delta{u_i},v_i+\Delta{v_i})\vert{i=1,\dots,M}\}$ by moving the M pixels according to the corresponding keypoint offsets. With the depth value $depth(c_i^{'})$ of the updated positions, the final PPCs can be determined in camera space. As shown in Figure \ref{fig:FtP}, the features of the PPCs $F_{pseu}\in\mathbb{R}^{M\times{C}}$ can be extracted from the frustum feature $F_T$ using the trilinear interpolation. Subsequently, to figure out the fusion between the features of PPCs and that of the raw points, the PPC $p_i^{'}=(u_i+\Delta{u_i},v_i+\Delta{v_i},depth(c_i^{'}))$ is re-projected to LiDAR space from the camera space by the transformation function $f_{re-proj}$ in KITTI
\begin{equation}
\begin{split}
    p_i^{pseu}&=(x^{'},y^{'},z^{'})=f_{re-proj}(p_i^{'})\\  &=T_{LiDAR\leftarrow{refer}}\cdot{T_{refer\leftarrow{camera}}\cdot{p_i^{'}}}
\end{split}
\end{equation}
where $p_i^{pseu}$ is the final coordinate of the $i^{th}$ PPC, $T_{refer\leftarrow{camera}}$ is the transformation matrix from the coordinate of color camera to the reference camera, and $T_{LiDAR\leftarrow{refer}}$ is the transformation matrix from the reference camera to LiDAR. To verify the effectiveness of Frustum-to-point Transition module, an alternative using M initial foreground points as PPCs is proposed and their features are extracted in the same way. In Section \ref{sec:Exp}, the comparison between two strategies on the KITTI dataset will be presented.
\par Overall, the multi-modal detection task is transformed into single-modal detection by using PPC instead of images to convey object information. A unified point-based representation helps to make subsequent interactions across multi-modal features easier.
	\begin{figure*}
		\centering
	\includegraphics[width=0.98\linewidth, height=6.5cm]{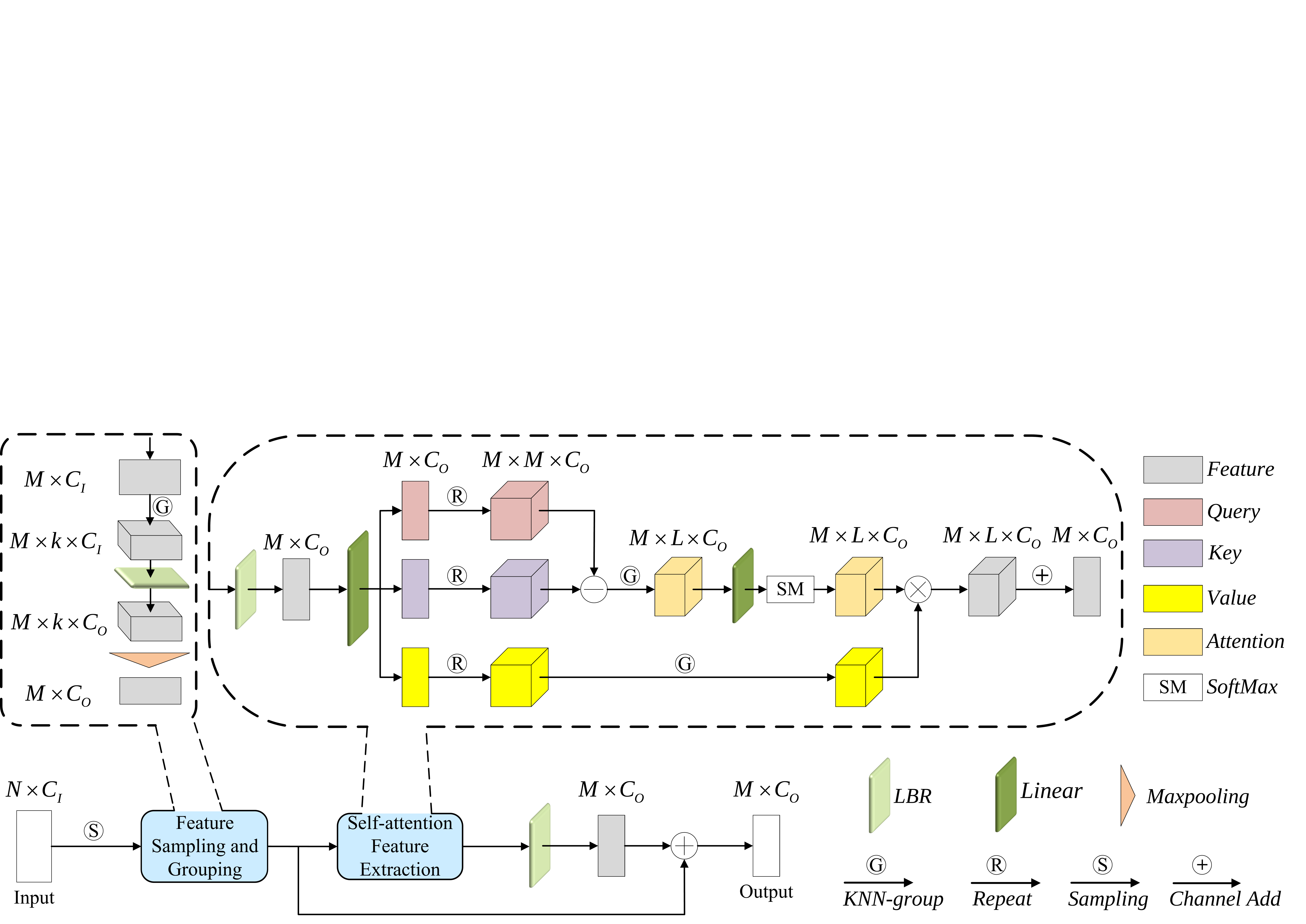}
		\vspace{-2mm}
		\caption{Illustration of PTD. The letters near the box indicate its shape, e.g., N$\times$C, N is the number of points and C is the dimension numbers of features. LBR combines Linear and BatchNorm layers and a ReLU function. KNN-group indicates the grouping operation with k-Nearest Neighbors algorithm.}
		\label{fig:PTD}
		\vspace{-3mm}
	\end{figure*}	
\subsection{Two-stream Feature Extraction Network}
\label{sec:3.2}
Multiple multi-modal methods \cite{DBLP:conf/cvpr/VoraLHB20,DBLP:journals/corr/abs-2007-08856,DBLP:journals/corr/abs-2112-11088} use a two-stream structure to process image and point cloud separately. Limited by the local receptive field of traditional building blocks, e.g., CNN and sparse convolution, these methods struggle to capture all useful feature relationships. In addition, feature alignment and fusion between image and point cloud are still the tricky problems.
\par Based on the unified point-based representation described above, a two-stream feature extraction network is developed to learn the features of point cloud and image at the point level. The two-stream network is mainly built on transformer for better feature learning and fusion. It has the inputs of the coordinates of point clouds $P_{raw}\in{\mathbb{R}^{N\times{3}}}$ and the coordinates of PPCs $P_{pseu}\in{\mathbb{R}^{M\times{3}}}$, and the corresponding features are $F_{raw}\in{\mathbb{R}^{N\times{C_1}}}$ and $F_{pseu}\in{\mathbb{R}^{M\times{C_2}}}$.\\
\textbf{Point Transition Down.} In the two-stream network, a stacked PTD encoder is responsible for iteratively extracting multilevel point representations. Based on recent attempts \cite{DBLP:journals/cvm/GuoCLMMH21,DBLP:journals/corr/abs-2012-09164} at object classification, PTD integrates the feature sampling and grouping, self-attention feature extraction and forward-feedback network into a separate module. In Figure \ref{fig:PTD}, PTD first subsamples M points from the input point $P_I$ (here $P_{raw}$ or $P_{pseu}$ can act as $P_I$) and use \emph{k}-NN algorithm to construct a neighbor embedding for each point. Then, an LBR (Linear layer, BatchNorm layer and ReLU function) operator and a max-pooling operator (MP) are used to encode local features as follows
\begin{equation}
    f_{local}(p)={\rm MP}({\rm LBR}(concat_{q\in{knn(p,P_I)}}\cdot f_I(q)))
\end{equation}
where $f_I(q)$ is the feature of point $q$ which belongs to the neighbor of point $p$, $knn(p,p_I)$ is k-nearest neighbors of point $p$ in $P_I$.
\par Next, we send the local feature $F_{local}\in{\mathbb{R}^{M\times{C_O}}}$ into self-attention feature extraction network to learn long-range dependencies of the features. The relationship between the \emph{query} (\emph{Q}), \emph{key} (\emph{K}), \emph{value} (\emph{V}) matrices and self-attention is as follows
\begin{gather}
    (Q, K, V)=\emph{LBR}(F_{local})\cdot{W_e}\notag\\
    Q, K, V\in{\mathbb{R}^{M\times{C_O}}}, W_e\in{\mathbb{R}^{C_O\times{3C_O}}}\notag\\  Q^{'}=rg(Q), K^{'}=rg(K), V^{'}=rg(V)\notag\\
    A^{'}=SM(\alpha(Q^{'}-K^{'}+\delta))
\end{gather}
where $W_e$ is the learnable weights of the linear layer and $rg(\cdot)$ represents repeat and grouping operation. $Q^{'}, K^{'}, V^{'}\in{\mathbb{R}^{M\times{L}\times{C_O}}}$ are the outputs after repeat and grouping operation related to the input \emph{Q}, \emph{K}, and \emph{V}. Furthermore, a position encoding defined as $\delta=\theta(p_i-p_j)$ is added to the attention, where $p_i$, $p_j$ are the coordinates of points $i$ and $j$. $\theta$ and $\alpha$ both consist of two linear layers and a ReLU function. Thereafter, the output of PTD could be derived as
\begin{equation}
F_O=\beta(sum(A^{'}\cdot(V^{'}+\delta)))+F_{local}
\end{equation}
where $sum(\cdot)$ represent the element-wise product, denotes channel-wise summation along the neighborhood axis, and $\beta$ is an LBR operator.
\par In the point cloud branch, the stacked PTD encoder (including four PTD modules) is used to learn point cloud features. In the PPC branch, the PPTD encoder adopts the same structure to extract image features.\\
\textbf{Point Transition Up.} In the point cloud branch, the stacked PTU decoder aims to restore the point cloud to its initial number and obtain the multi-scale features for proposal generation. PTU can be easily constructed by replacing the feature sampling and grouping in PTD with the inverse distance-weighted average operation, in the meanwhile, keeping the other structures intact. The inverse distance-weighted average operation is proposed as the skip connection in PointNet++ \cite{DBLP:conf/nips/QiYSG17}
\begin{equation}
    f_{int}(p_i)=\frac{\sum\nolimits_{j=1}^{k}w_j\cdot f_e(p_j)}{\sum\nolimits_{j=1}^{k}w_j}
\end{equation}
where $w_j=\frac{1}{d(p_i,p_j)^p},j=1,\dots,k$, $p_i$ is the coordinate of the interpolated point, $p_j$ is the coordinate of the neighboring point of $p_i$, $d(\cdot,\cdot)$ denotes the Eu-clidean distance between two points, and $f_{int}(p_i)$ denotes the interpolated features of $p_i$. Let $p=2$, $k=3$ be the same settings in PointNet++ \cite{DBLP:conf/nips/QiYSG17}, then, the interpolated features are added by the skip connection features as
\begin{equation}
F_I^{n}=F_{int}^{n}+{\rm LBR}(F_{skip}^{n}),n=1,2,3,4
\end{equation}
	\begin{figure}
		\centering
	\includegraphics[width=0.88\columnwidth]{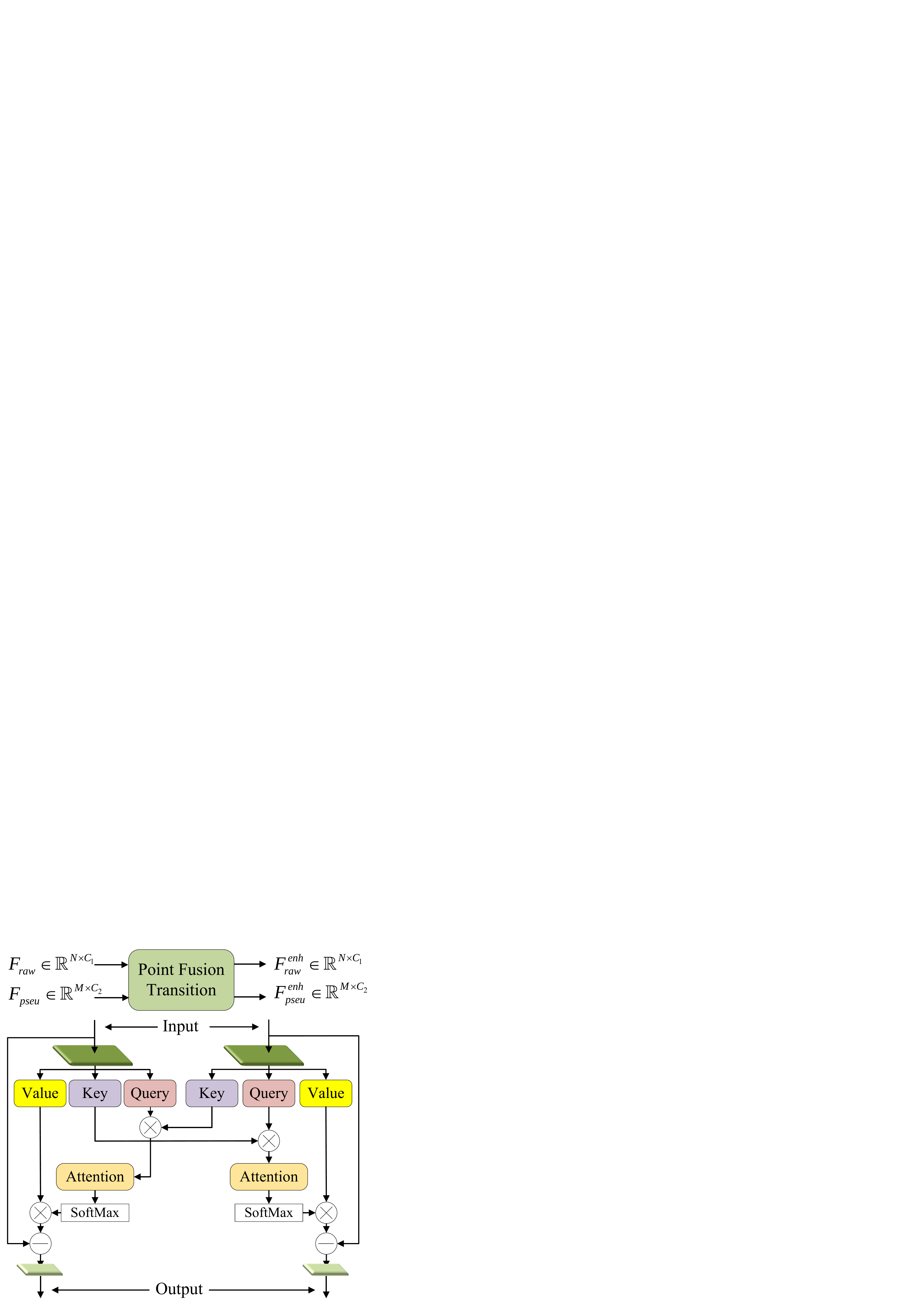}
		\vspace{-3mm}
		\caption{Illustration of PFT. It is employed to perform bi-directional information exchange between two point-based features from different modalities.}
		\vspace{-5mm}
		\label{fig:PFT}
	\end{figure}
where $F_{skip}^{n}$ is the $n$-$th$ output of the PTD and $F_{int}^{n}$ is the interpolated feature of the $n$-$th$ PTU. $F_I^{n}$ is used as the input of the remaining structure of the $n$-$th$ PTU. On the contrary, in the PPC branch, a stacked FP decoder with four FP layers is used to recover the initial PPCs. Since the position of the PPC is defined on the object keypoint, the distribution of the PPC is more focused on the object than the point cloud directly sampled from the LiDAR. Meanwhile, considering the large memory and time overhead of PTU itself, the FP layer is selected to handle the PPC that does not require a large receptive field.\\
\textbf{Point Fusion Transition.} According to the above introduction, the stacked PTD encoders of the two branches simultaneously extract point-based features layer-by-layer. However, the features from the point cloud branch lack the semantic and textural information of the object, and the features from the PPC branch lack the geometry information for locating the object. Moreover, both the point cloud provided by LiDAR and the image-generated PPC are inevitably contaminated by noise. To address these problems, a dual input and dual output PFT module is designed for feature fusion in Figure \ref{fig:PFT}. PFT fuses two input features based on cross-attention and produces two enhanced features as the inputs to the next level. Finally, an additional PFT is used to fuse the outputs of the two branches (see Figure \ref{fig:backbone}) to obtain the final point representations.
\par PFT module is also based on transformer and the $Q$, $K$, and $V$ matrices are generated separately for the two inputs
\begin{gather}(Q_{raw},K_{raw},V_{raw})=F_{raw}\cdot{W_{raw}}\notag\\
(Q_{pseu},K_{pseu},V_{pseu})=F_{pseu}\cdot{W_{pseu}}\notag\\
Q_{raw},K_{raw},V_{raw}\in{\mathbb{R}^{N\times{C_e}}}\notag\\Q_{pseu},K_{pseu},V_{pseu}\in{\mathbb{R}^{M\times{C_e}}}\notag\\ W_{raw}\in{\mathbb{R}^{C_1\times{3C_e}}},W_{pseu}\in{\mathbb{R}^{C_2\times{3C_e}}}
\end{gather}
where $W_{raw}$ and $W_{pseu}$ are both learnable weights. Then, the cross-attention for each data is defined as
\begin{gather}
A_{raw}=SM(\sigma(K_{raw}\cdot{Q_{pseu}^T}))\notag\\
A_{pseu}=SM(\varepsilon(K_{pseu}\cdot{Q_{raw}^T}))\notag\\
A_{raw}\in{\mathbb{R}^{N\times{M}}},A_{pseu}\in{\mathbb{R}^{M\times{N}}}
\end{gather}
where $\sigma$ and $\varepsilon$ both comprise two linear layers and a ReLU function. Here, we multiply the $K$ matrix of one modality by the $Q$ matrix of the other modality to generate cross-attention. It differs from the way computed in PTD. This practice is inspired by HVPR \cite{DBLP:conf/cvpr/NohLH21} which takes voxel-based features as queries and computes matching probabilities between the voxel-based features and the memory items through dot product. In Section \ref{sec:Exp}, we conduct ablation experiments to compare the effects of different attention calculation ways. Finally, the enhanced features as the outputs of PFT can be expressed as
\begin{gather}
F_{raw}^{enh}={\rm LBR}(F_{raw}-(A_{raw}\cdot{V_{pseu}}))\notag\\
F_{pseu}^{enh}={\rm LBR}(F_{pseu}-(A_{pseu}\cdot{V_{raw}}))
\end{gather}
\par It is worth mentioning that Zhang et al \cite{DBLP:journals/corr/abs-2204-00325} propose a similar structure to PFT. However, they have the limitations that the information can only flow from the image domain to the point domain. In contrast, PFT conducts bidirectional information exchange which provides semantic information for point cloud and geometry information for PPC.

\subsection{RPN and Refinement network}
The two-stream feature extraction network described in subsection \ref{sec:3.2} aims to learn expressive features for every LiDAR point. After that, the features will be sent to the RPN to generate proposals. The RPN has a classification head and a regression head. To obtain high-quality proposals, 3D votes are computed as suggested by ImVoteNet \cite{DBLP:conf/cvpr/QiCLG20}, since 3D votes can help narrow down the search space from point to proposal center. The votes are then concatenated with the output of the two-stream network. Finally, the enhanced features are fed into the RPN. After acquiring the proposals, non-maximum suppression (NMS) is applied to eliminate redundant proposals. The remaining proposals are sent to the refinement network for generating bounding boxes. In the experiments, two refinement strategies, including point cloud region pooling and Roi-aware point cloud feature pooling are adopted, as proposed by PointRCNN \cite{DBLP:conf/cvpr/ShiWL19} and Part-A2 \cite{DBLP:journals/pami/ShiWSWL21}, respectively. Actually, our PTA-Det can be plugged into most point-based detectors as multi-modal detectors.
\begin{table*}[t]
		\caption{Comparison with popular 3D object detection methods on the KITTI validation split. The available results are evaluated by mAP for each category. ‘L’ and ‘R’ stand for LiDAR and image, respectively. $\star$ indicates no data augmentation technique is applied. PTA-Det-1 and PTA-Det-2 represent our method adopts the refinement network proposed in PointRCNN and Part-$A^2$, respectively. The optimal results are marked in bold font.}
		\vspace{-2mm}
		\centering
		\begin{adjustbox}{width=0.96\textwidth}
			\begin{tabular}{c|c|cccc|cccc|cccc|c}
				\hline
				\multirow{2}{*}{Method} &
                \multirow{2}{*}{Modality} & 
				\multicolumn{4}{c|}{Car } &
				\multicolumn{4}{c|}{Pedestrian }  &
				\multicolumn{4}{c|}{Cyclist } &
                \multirow{2}{*}{3D mAP (\%)}\\
				
				\cline{3-14}
				
				{}  & {} & Easy & Moderate  & Hard & mAP & Easy & Moderate  & Hard & mAP  & Easy & Moderate  & Hard  & mAP & {}\\ \hline\hline
                SECOND\cite{DBLP:journals/sensors/YanML18} &L &88.61 &78.62	&77.22 &81.48 &56.55 &52.98	&47.73	&52.42	&80.58	&\textbf{67.15}	&\textbf{63.10}	&\textbf{70.28}	&68.06\\
                ${\rm PointRCNN}^{\star}$\cite{DBLP:conf/cvpr/ShiWL19} &L &89.41	&78.10	&75.51	&81.01	&\textbf{70.39}	&\textbf{60.41}	&\textbf{51.48}	&\textbf{60.79}	&\textbf{81.81}	&58.10	&53.86	&64.59	&\textbf{68.80}\\
                Part-$A^{2\star}$\cite{DBLP:journals/pami/ShiWSWL21} &L &85.28 &74.22	&69.85	&76.45	&53.24	&46.80	&41.06	&45.50	&66.51	&41.78	&39.30	&66.54	&63.99\\ 
                SE-SSD\cite{DBLP:conf/cvpr/ZhengTJF21} &L &\textbf{90.21} &\textbf{86.25}	&\textbf{79.22}	&\textbf{85.23}	&-	&-	&-	&-	&-	&-	&-	&-	&-\\ \hline \hline
                MV3D\cite{DBLP:journals/corr/ChenMWLX16} &L+R &71.29 &62.68	&56.56	&63.51	&-	&-	&-	&-	&-	&-	&-	&-	&-\\
                AVOD-FPN\cite{DBLP:conf/iros/KuMLHW18} &L+R &84.41 &74.44	&68.65	&75.83	&-	&58.80	&-	&-	&-	&49.70	&-	&-	&-\\
                F-PointNet\cite{DBLP:conf/cvpr/QiLWSG18} &L+R	&83.76 &70.92 &63.65 &72.78	&\textbf{70.00}	&\textbf{61.32}	&53.59	&\textbf{61.64}	&77.15	&56.49	&53.37	&62.34	&65.58\\
                SIFRNet\cite{DBLP:conf/aaai/ZhaoLHH19} &L+R &85.62	&72.05 &64.19 &73.95 &69.35 &60.85 &52.95 &61.05 &80.97 &60.34 &56.69	&65.97	&66.99\\
                EPNET\cite{DBLP:journals/corr/abs-2007-08856} &L+R	&\textbf{88.76}	&\textbf{78.65}	&\textbf{78.32}	&\textbf{81.91}	&66.74	&59.29	&\textbf{54.82}	&60.28	&\textbf{83.88}	&\textbf{65.50}	&\textbf{62.70}	&\textbf{70.69}	&\textbf{70.96}\\ \hline \hline
                PTA-Det-1 &L+R &\textbf{86.31}	&\textbf{77.06}	&\textbf{70.28}	&\textbf{77.88}	&61.77	&51.84	&46.98	&53.53	&70.61	&49.02	&45.54	&55.06	&62.16\\
                PTA-Det-2 &L+R &84.72	&74.45	&69.86	&76.34	&60.84	&52.48	&45.11	&52.81	&72.43	&49.17	&46.75	&56.12	&61.76\\
                \hline 
		\end{tabular}
	\end{adjustbox}
	\label{tbl:mainresult}
	\vspace{-2mm}
	\end{table*}
\subsection{Overall Loss Function}
The model is optimized by a multi-task loss which can be formulated as
\begin{equation}
L_{total}=\lambda_{depth}L_{depth}+\lambda_{rpn}L_{rpn}+\lambda_{rcnn}L_{rcnn}
\end{equation}
where the $L_{depth}$ denotes the loss of depth prediction for generating the PPCs in Pseudo Point Cloud Generation network. $L_{rpn}$ is the loss of the two-stream feature extraction network to generate the proposal. $L_{rcnn}$ is the loss of the refinement network. $\lambda_{depth}$, $\lambda_{rpn}$, and $\lambda_{rcnn}$  are fixed loss weighting factors. $L_{depth}$ can be computed as
\begin{equation}
L_{depth}=L_{bin}+\lambda_{1}L_{res}
\end{equation}
where $\lambda_{1}$ is the balance weight for depth residual with the setting of $\lambda_{1}=10$. $L_{bin}$ and $L_{res}$ are defined as
\begin{gather}
L_{bin}=\frac{1}{N}\sum\nolimits_{i=1}^{N}{\rm FL}(D_{bin}(u_i,v_i),D_{{gt}\_{bin}}^{i})\\
L_{res}= \frac{1}{N}\sum\nolimits_{i=1}^{N}{\rm SML}(D_{res}(u_i,v_i),D_{{gt}\_{res}}^{i})
\end{gather}
where {\rm FL} denotes focal loss \cite{DBLP:journals/pami/LinGGHD20} and {\rm SML} is $Smooth-{\rm L1}$ loss. $D_{gt\_{bin}}^i$ and $D_{gt\_{res}}^i$ denote bin-based index and normalized residual value of the $i^{th}$ foreground point’s depth. $D_{bin}(u_i,v_i)$ and $D_{res}(u_i,v_i)$ have been introduced in section \ref{sec:3.1}, and the focal loss is adopted in $L_{bin}$ with the setting of $\alpha=0.25$ and $\lambda=2.0$. $L_{rpn}$ consists of a classification loss and a regression loss as
\begin{equation}
L_{rpn}= L_{cls}+\lambda_2{L_{reg}}
\end{equation}
with
\begin{equation}
L_{cls}= -\alpha_t(1-c_t)^\gamma{log(c_t)}
\end{equation}
and
\begin{equation}
\begin{split}
c_t=\begin{cases} c &if\;p\;is\;the\;foreground\;point\;\\ 
1-c &otherwise\end{cases}
\end{split}
\end{equation}
\begin{equation}
L_{reg}=\sum\nolimits_{u\in(x,y,z,l,w,h,sin\theta,cos\theta)}{\rm SML} (\hat{res_u},res_u)
\end{equation}
where $\lambda_2$ is the balance weight, $c$ denotes the classification confidence for the point $p$, and $L_{cls}$ is supervised by focal loss. $\hat{res_u}$ and $res_u$ are the predicted residuals and residual labels of the foreground point. $Smooth$-${\rm L1}$ loss is used to regress the offsets of the location, size, and direction. The loss of the refinement network is the same as that of PointRCNN \cite{DBLP:conf/cvpr/ShiWL19} or Part-$A^2$ \cite{DBLP:journals/pami/ShiWSWL21}.
 	\begin{table*}[t]
		\caption{Detection performance of PointRCNN on the validation set under different sampling strategies.}
			\vspace{-2mm}
		\centering
		\begin{adjustbox}{width=0.60\textwidth}
			\begin{tabular}{c|c|ccc|c}
				\hline
				\multicolumn{2}{c|}{Sampling Strategy } &
				\multicolumn{4}{c}{3D Object Detection (\%)}  \\ \hline
				Input Point  &Group Size	&Car	&Cyc.	&Ped.	&mAP\\ \hline\hline
				16384	&4096\text{--}1024\text{--}256\text{--}64	&81.01	&64.59	&60.79	&68.80\\ 
                8000	&2000\text{--}1000\text{--}500\text{--}250	&76.55	&48.28	&52.86	&59.23\\ 
                4000	&2000\text{--}1000\text{--}500\text{--}250	&68.65	&37.27	&45.09	&50.34\\ 
                1600	&800\text{--}400\text{--}200\text{--}100	&38.28	&31.05	&2.36	&23.90\\ \hline
		\end{tabular}
	\end{adjustbox}
	\label{tb2:pointrcnn}
	\vspace{-2mm}
	\end{table*}
 
\section{Experiment}
\label{sec:Exp}
The model is evaluated on KITTI, a commonly used benchmark dataset for 3D object detection. PTA-Det is built on the basis of the OpenPCDet \cite{openpcdet2020} which is an open-source project for 3D object detection.

\subsection{Dataset and evaluation metric} The KITTI dataset consists of 7481 training samples and 7518 test samples, focusing on the categories of car, pedestrian and cyclist. Following the investigations \cite{DBLP:journals/corr/ChenMWLX16,DBLP:conf/cvpr/QiLWSG18}, the original training samples are further separated into a training set (3712 frames) and a validation set (3769 frames). The Average Precision (AP) is calculated using 40 recall positions as the validation metric according to Geiger et al \cite{DBLP:conf/cvpr/GeigerLU12}. All the objects are classified into easy, moderate, and hard levels based on their sizes, occlusion, and truncation. In the experiments, the results on the validation set are reported for all difficulty levels.

\subsection{Implementation Details} \noindent\textbf{Network settings.}As a multi-modal 3D object detection method, LiDAR points, RGB image, and camera calibration matrices are taken as inputs. We assume that the 3D scene is constrained to [(0,70.4),(-40,40),(-3,1)] meters along the X (forward), Y (left), and Z (up) axes in the LiDAR coordinate, respectively. During depth prediction, the depth range [0,70.4]is discretized into 80 bins. Unlike LiDAR coordinate, the camera coordinate is set along the X (left), Y (down), and Z (forward) axes. The transformation between two coordinates can be achieved by a calibration matrix. 
\par For each 3D scene, 16000 lidar points and images with a resolution of 1280$\times$384 are used as initial inputs to the model. In the Pseudo Point Cloud Generation Network, Mask-RCNN implemented by detectron2 \cite{wu2019detectron2} is used to generate the foreground mask, and 4096 foreground points are selected through the mask to guide the depth prediction. In a scene where the number of foreground points is less than 4096, the remaining points are randomly selected from the background points. Then, 1600 points are further sampled from the foreground points as the input to the point cloud branch, where the stacked PTD encoder has the point numbers set to 800, 400, 200, 100, respectively. By contrast, the Pseudo Point Cloud Generation Network produces 480 PPCs for the PPC branch, and the PPTD encoder iteratively extracts the features of PPCs, whose numbers are 240, 120, 60, 30, respectively.\\
\textbf{Training scheme.} After generating proposals in RPN, redundant proposals are eliminated using NMS. The thresholds are set to 0.8 and 0.85 in the training and testing stages. In the refinement network, we utilize the IoU between the proposal and the ground truth to distinguish between positive and negative proposals. Following PointRCNN \cite{DBLP:conf/cvpr/ShiWL19}, different thresholds are selected for classification and regression. Specifically, the proposals with IoU scores higher than 0.6 are considered positive samples for classification. In contrast, the proposals with IoU scores lower than 0.45 are considered negative samples. The proposals with IoU scores higher than 0.55 are used to calculate the regression loss.
\par We train the model with a batch size of 2 for 80 epochs and adopt the Adaptive Moment Estimator (Adam) optimizer with an initial learning rate, weight decay, and momentum at 0.01, 0.01, and 0.9, respectively. All experiments are conducted on two RTX 3090 GPUs using the deep learning framework PyTorch \cite{2017Automatic}. It is worth noting we do not use any data augmentation techniques during training. 

\subsection{Main Results}
PTA-Det is compared with several LiDAR-only and multi-modal 3D object detection methods,and the results are summarized in Table \ref{tbl:mainresult}. Two versions of PTA-Det are given, one of which adopts the point cloud region pooling strategy and the other adopts the ROI-aware point cloud pooling strategy. The mAP of the former is 1.54\% and 0.72\% higher than that of the latter for the car and pedestrian categories, respectively. But in the cyclist category, the latter outperforms the former by 1.06\%. The results show that the refinement strategy employed by the former is sufficient to accurately localize the object when image features are used as a complement. Although the latter can better capture the point cloud distribution in the proposals, its advantage is not obvious in our model.
\par Table \ref{tbl:mainresult} shows that PTA-Det exhibits better performance than a variety of previous multimodal methods by about 2\% to 14\% mAP in the car category. However, the current PTA-Det shows less mAP in the pedestrian and cyclist categories. The reason is that, considering the large memory overhead in the transformer, the number of point clouds input to the model is reduced. This makes the points on the surface of small objects more sparse, which in turn leads to poor detection performance of the model on small objects.
\par To reveal the reason for the degradation, we study the performance of PointRCNN \cite{DBLP:conf/cvpr/ShiWL19} under different sampling strategies, as summarized in Table \ref{tb2:pointrcnn}. PointRCNN \cite{DBLP:conf/cvpr/ShiWL19} abstracts a set of point representations using iterative set-abstraction (SA) blocks. In the default case, it samples 16384 points from the scene as input, and uses four SA layers in sequence to sample the points with group sizes of 4096, 1024, 256 and 64 respectively. Three different sampling strategies are also presented, the last of which uses the same sampling strategy as PTA-Det.
\par As shown in Table \ref{tb2:pointrcnn}, the mAP of PointRCNN decreases significantly in all categories as the number of input points decreases. Comparing the fourth strategy with PTA-Det-1, it is proved that PTA-Det can reach a higher accuracy than PointRCNN with the same number of points. These investigations support our conjecture above on the reason for the poor performance of PTA-Det when detecting small objects. 
\par
\begin{figure*}
		\centering
	\includegraphics[width=0.88\linewidth, height=6.5cm]{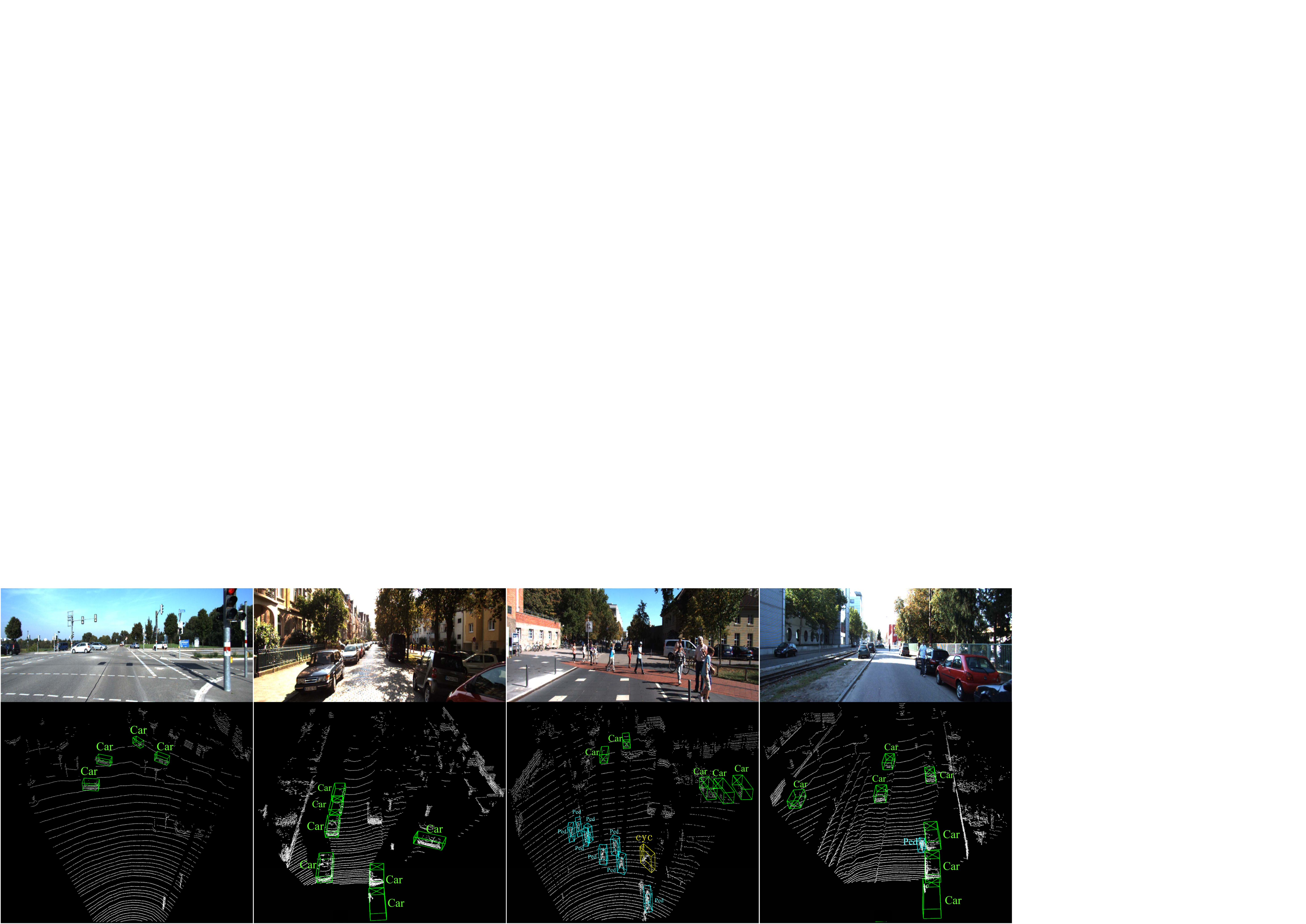}
		\vspace{-2mm}
		\caption{Visualized results by PTA-Det on KITTI test set. Green/turquoise/yellow bounding box indicates car/pedestrian/cyclist category, respectively. The scenes are arranged from left to right. The first and fourth show that PTA-Det performs well against distant objects. The second successfully detects all cars and excludes the object of van category. From the third scene, we observe that PTA-Det performs a little worse with multiple pedestrians and overlooks a car due to occlusion.}
		\label{fig:test-4}
		\vspace{-3mm}
	\end{figure*}

 PTA-Det does not adopt data augmentation technique because it is complex and rarely considered for multi-modal scenarios. For a fair comparison, we also provide the results of PointRCNN and Part-A2 without data augmentation in Table \ref{tbl:mainresult}. PTA-Det performs similarly to them in the car category. Finally, we made several qualitative investigations to illustrate the effectiveness of PTA-Det, as visualized in Figure \ref{fig:test-4}.

\subsection{Ablation Studies}
We perform ablation studies to evaluate the influence of each module or strategy on final results, including PTD, PTU, PFT, two selection strategies for PPC, and the calculation way of attention in PFT. For comparison, a baseline two-stream structure is designed to replace the two-stream network of PTA-Det. It also contains a point cloud branch and a PPC branch, both of which are built with PointNet++ block and FP layer. In addition, the interaction between the two branches is achieved through simple feature concatenation at multiple levels. Other network structures and parameter settings remain unchanged in the baseline model. Both PTA-Det and the baseline selects the Roi-aware point cloud feature pooling strategy in the refinement network. To trade off the speed and accuracy of PTA-Det, we choose ResNet-50 as the image backbone and follow the research of Pan et al \cite{DBLP:conf/cvpr/PanXSLH21} to use the computational cost reduction strategy in PFT.\\
\textbf{Effects of PTD and PTU modules.} From the first three rows in Table \ref{tb3:3Presult}, the baseline obtains a 43.25\% mAP. After replacing all PointNet++ blocks with PTD, the mAP is improved by 5.59\%. If all FP layers in the point cloud branch are replaced with PTU, the mAP is improved by 11.13\%. When we use the two strategies together in the $5^{th}$ row, the mAP is improved by 15.52\% relative to that of the baseline. The improvements are attributed to the self-attention mechanism, which can aggregate long-distance dependencies better than Pointnet++ \cite{DBLP:conf/nips/QiYSG17} blocks and FP layers \cite{DBLP:conf/nips/QiYSG17}.\\ 
 
 	\begin{table*}[t]
		\caption{Contributions of PTD/PTU/PFT modules to PTA-Det.}
			\vspace{-2mm}
		\centering
        \tiny
	\begin{adjustbox}{width=0.65\textwidth}
			 \begin{tabular}{cccc|ccc|c}
				\hline
				\multicolumn{4}{c|}{Component} &
				\multicolumn{4}{c}{3D Object Detection (\%)}  \\ \hline
    Baseline	&PTD	&PTU	&PFT	&Car	&Cyc.	&Ped.	&mAP\\ \hline\hline
    $\checkmark$	&{} &{} &{}			&68.94	&32.84	&27.96	&43.25\\ 
    {}	&$\checkmark$ &{} &{}			&65.77	&42.62	&38.12	&48.84\\ 
    {} &{}		&$\checkmark$	&{}	&72.66	&43.38	&47.11	&54.38\\
    {} &{} &{}	&$\checkmark$	&66.74	&32.44	&28.20	&42.46\\
    {} &$\checkmark$  &$\checkmark$  &{}		&75.03	&51.73	&49.56	&58.77\\ 
    {}	&$\checkmark$	&$\checkmark$	&$\checkmark$	&75.98	&52.00	&52.41	&60.13\\ \hline
		\end{tabular}
	\end{adjustbox}

	\label{tb3:3Presult}
	\vspace{-2mm}
	\end{table*}
 
 	\begin{table}[t]
		\caption{ Comparison of fusion structure in PFT to PTA-Det.}
		\vspace{-2mm}
		\centering
		\begin{adjustbox}{width=0.99\columnwidth}
			\begin{tabular}{c|ccc|c}
				\hline
				\multirow{2}{*}{Integrating Scheme} &
				\multicolumn{4}{c}{3D Object Detection (\%)} \\
				\cline{2-5}
                {} &Car	&Cyc. &Ped.	&mAP\\ \hline\hline
                ${\rm PFT}_c$ &75.22	&48.26	&47.52	&57.00\\
                ${\rm PFT}_+$ &75.41	&51.69	&50.13	&59.08\\
                ${\rm PFT}_\_$ &75.98	&52.00	&52.41	&60.13\\\hline
			\end{tabular}
		\end{adjustbox}
		\label{tb4:PFTresult}
		\vspace{-3mm}
	\end{table}
\noindent\textbf{Effects of PFT and Fusion Operation in PFT.}
Table \ref{tb3:3Presult} also shows that after introducing PFT into the baseline network, although the mAP of the pedestrian category is improved by 0.24\%, the performance of the other categories becomes worse. This is because it is difficult for PFT to directly compute the attention of multi-scale features from SA blocks in Pointnet++. However, as can be found in the $5^{th}$ and $6^{th}$ rows, PFT can clearly improve the mAP if accompanied by PTD and PTU. A total improvement up to 16.88\% has been realized when PTD, PTU and PFT are used simultaneously.
\par To analyze the impact of the structure of the PFT on the results, three schemes have been studied in Table \ref{tb4:PFTresult}. In PFT, by default, the cross-modal features of each modality are subtracted from its input features, and attached with an LBR as the output. This module is denoted as ${\rm PFT}_\_$. We then replace subtraction with summation and concatenation, denoted as ${\rm PFT}_+$ and ${\rm PFT}_c$, respectively. As in Table \ref{tb4:PFTresult}, ${\rm PFT}_\_$ exhibits better performance than the two schemes. The improvements are +3.13\% and +1.05\% on mAP, showing the advantage of subtraction in PFT.\\
 	\begin{table}[t]
		\caption{Comparison of calculation ways of attention. The three consecutive symbols represent the way of attention computation used in PTD, PTU, and PFT, where \text{--} stands for subtraction, × for multiplication.}
		\vspace{-2mm}
		\centering
		\begin{adjustbox}{width=0.99\columnwidth}
			\begin{tabular}{c|ccc|c}
				\hline
				\multirow{2}{*}{Calculation of Attention} &
				\multicolumn{4}{c}{3D Object Detection (\%)} \\
				\cline{2-5}
                {} &Car	&Cyc. &Ped.	&mAP\\ \hline\hline
                \text{--} \text{--} \text{--}	&72.47	&50.26	&43.58	&55.44\\
               × × ×	&70.98	&45.27	&43.03	&53.09\\
               × × \text{--}	&69.34	&48.22	&39.79	&52.45\\
               \text{--} \text{--} ×	&75.98	&52.00	&52.41	&60.13\\ \hline
			\end{tabular}
		\end{adjustbox}
		\label{tb5:ATTresult}
		\vspace{-3mm}
	\end{table}
\textbf{Influence of the Calculation Way of Attention in PTD, PTU and PFT.} In PTD and PTU, the subtraction between the query matrix and the key matrix is used to compute the self-attention of the point features, while the multiplication between the two matrices is used in PFT to compute the cross-modal attention. In order to investigate the impact of the two attention computation ways on mAP, we combine the PTD, PTU, and PFT modules using subtraction or multiplication operation to form four different schemes for comparison. Table \ref{tb5:ATTresult} shows that the fourth way has 4.69\% to 7.68\% higher mAP than the other three ways. The subtraction between point-based features helps capture the relationship between different features in the same dimension, since it provides the attention computation for each channel. This is crucial for PTD and PTU to obtain intra-modal features. In contrast, inter-modal features vary greatly and a larger perspective is required to capture their relationships. Multiplication can produce a scalar to measure the distance between features across channels. Thus, multiplication is more suitable for computing cross-attention than subtraction for PFT.\\
 	\begin{table}[t]
		\caption{Comparison of sampling strategies of Pseudo Point Cloud Generation network to PTA-Det.}
		\vspace{-2mm}
		\centering
		\begin{adjustbox}{width=0.99\columnwidth}
			\begin{tabular}{cc|ccc|c}
				\hline
				\multicolumn{2}{c}{Sampling Strategy} &
				\multicolumn{4}{c}{3D Object Detection (\%)} \\
				\hline
                FPS & KPS & Car & Cyc. &Ped. &mAP\\\hline\hline
                $\checkmark$ &{}	&73.83	&53.76	&48.79	&58.79\\
	        {}  &$\checkmark$	&75.98	&52.00	&52.41	&60.13\\\hline
			\end{tabular}
		\end{adjustbox}
		\label{tb6:ssresult}
		\vspace{-3mm}
	\end{table}
\textbf{Effect of Sampling Strategy in Pseudo Point Cloud Generation Network.} In section \ref{sec:3.1}, we mentioned two strategies to obtain PPCs: 1) apply FPS algorithm to sample the foreground points and the sampled points are used as the final PPCs, which is denoted as FPS; 2) apply the keypoint sampling strategy based on 2D keypoint offset, which is denoted as KPS. Table \ref{tb6:ssresult} shows that KPS performs better. The results show that using the object keypoints as PPCs can provide more information about the object than PPCs directly sampled from the foreground points.
\section{Conclusion}
\label{sec:Con}
In this paper, a method named PTA-Det is proposed, which uses pseudo points as an intermediate modality between the image and the point cloud to solve the multi-modal 3D object detection problem. The pseudo points generated by a Pseudo Point Cloud Generation network not only contain representative semantic and textural information, but also compensate for the missing information of the object. The generated PPC and point cloud are then fed into a two-stream transformer-based feature extraction network to learn intra-modal and inter-modal features. PTA-Det aims to explore a more reasonable fusion method for multi-modal 3D object detection and form a plug-and-play module that can be combined with LiDAR-only methods. Extensive experiments are conducted on the KITTI dataset and competitive results are given. PTA-Det shows better performance on car category with a mAP of 77.88\%, while less effective on cyclist and pedestrian categories. With the same input points, our method can achieve better accuracy than the LiDAR-only methods. The experimental results indicate PTA-Det could be a robust approach for 3D object detection in autonomous driving, and many other applications.
	
{\small
	\bibliographystyle{ieee_fullname}
	\bibliography{pta-det}

\begin{thebibliography}{10}\itemsep=-1pt

\bibitem{DBLP:conf/cvpr/BaiHZHCFT22}
Xuyang Bai, Zeyu Hu, Xinge Zhu, Qingqiu Huang, Yilun Chen, Hongbo Fu, and
  Chiew{-}Lan Tai.
\newblock Transfusion: Robust lidar-camera fusion for 3d object detection with
  transformers.
\newblock In {\em {IEEE/CVF} Conference on Computer Vision and Pattern
  Recognition, {CVPR} 2022, New Orleans, LA, USA, June 18-24, 2022}, pages
  1080--1089. {IEEE}, 2022.

\bibitem{CHEN202223}
Wenyu Chen, Peixuan Li, and Huaici Zhao.
\newblock Msl3d: 3d object detection from monocular, stereo and point cloud for
  autonomous driving.
\newblock {\em Neurocomputing}, 494:23--32, 2022.

\bibitem{DBLP:journals/corr/ChenMWLX16}
Xiaozhi Chen, Huimin Ma, Ji Wan, Bo Li, and Tian Xia.
\newblock Multi-view 3d object detection network for autonomous driving.
\newblock {\em CoRR}, abs/1611.07759, 2016.

\bibitem{DBLP:conf/iccv/DaiQXLZHW17}
Jifeng Dai, Haozhi Qi, Yuwen Xiong, Yi Li, Guodong Zhang, Han Hu, and Yichen
  Wei.
\newblock Deformable convolutional networks.
\newblock In {\em {IEEE} International Conference on Computer Vision, {ICCV}
  2017, Venice, Italy, October 22-29, 2017}, pages 764--773. {IEEE} Computer
  Society, 2017.

\bibitem{DBLP:conf/iclr/DosovitskiyB0WZ21}
Alexey Dosovitskiy, Lucas Beyer, Alexander Kolesnikov, Dirk Weissenborn,
  Xiaohua Zhai, Thomas Unterthiner, Mostafa Dehghani, Matthias Minderer, Georg
  Heigold, Sylvain Gelly, Jakob Uszkoreit, and Neil Houlsby.
\newblock An image is worth 16x16 words: Transformers for image recognition at
  scale.
\newblock In {\em 9th International Conference on Learning Representations,
  {ICLR} 2021, Virtual Event, Austria, May 3-7, 2021}. OpenReview.net, 2021.

\bibitem{DBLP:journals/ijrr/GeigerLSU13}
Andreas Geiger, Philip Lenz, Christoph Stiller, and Raquel Urtasun.
\newblock Vision meets robotics: The {KITTI} dataset.
\newblock {\em Int. J. Robotics Res.}, 32(11):1231--1237, 2013.

\bibitem{DBLP:conf/cvpr/GeigerLU12}
Andreas Geiger, Philip Lenz, and Raquel Urtasun.
\newblock Are we ready for autonomous driving? the {KITTI} vision benchmark
  suite.
\newblock In {\em 2012 {IEEE} Conference on Computer Vision and Pattern
  Recognition, Providence, RI, USA, June 16-21, 2012}, pages 3354--3361. {IEEE}
  Computer Society, 2012.

\bibitem{DBLP:journals/cvm/GuoCLMMH21}
Meng{-}Hao Guo, Junxiong Cai, Zheng{-}Ning Liu, Tai{-}Jiang Mu, Ralph~R.
  Martin, and Shi{-}Min Hu.
\newblock {PCT:} point cloud transformer.
\newblock {\em Comput. Vis. Media}, 7(2):187--199, 2021.

\bibitem{DBLP:conf/iccv/HeGDG17}
Kaiming He, Georgia Gkioxari, Piotr Doll{\'{a}}r, and Ross~B. Girshick.
\newblock Mask {R-CNN}.
\newblock In {\em {IEEE} International Conference on Computer Vision, {ICCV}
  2017, Venice, Italy, October 22-29, 2017}, pages 2980--2988. {IEEE} Computer
  Society, 2017.

\bibitem{DBLP:conf/cvpr/HeZRS16}
Kaiming He, Xiangyu Zhang, Shaoqing Ren, and Jian Sun.
\newblock Deep residual learning for image recognition.
\newblock In {\em 2016 {IEEE} Conference on Computer Vision and Pattern
  Recognition, {CVPR} 2016, Las Vegas, NV, USA, June 27-30, 2016}, pages
  770--778. {IEEE} Computer Society, 2016.

\bibitem{DBLP:journals/corr/abs-2007-08856}
Tengteng Huang, Zhe Liu, Xiwu Chen, and Xiang Bai.
\newblock Epnet: Enhancing point features with image semantics for 3d object
  detection.
\newblock {\em CoRR}, abs/2007.08856, 2020.

\bibitem{NIPS2012_c399862d}
Alex Krizhevsky, Ilya Sutskever, and Geoffrey~E Hinton.
\newblock Imagenet classification with deep convolutional neural networks.
\newblock In F. Pereira, C.J. Burges, L. Bottou, and K.Q. Weinberger, editors,
  {\em Advances in Neural Information Processing Systems}, volume~25. Curran
  Associates, Inc., 2012.

\bibitem{DBLP:conf/iros/KuMLHW18}
Jason Ku, Melissa Mozifian, Jungwook Lee, Ali Harakeh, and Steven~L. Waslander.
\newblock Joint 3d proposal generation and object detection from view
  aggregation.
\newblock In {\em 2018 {IEEE/RSJ} International Conference on Intelligent
  Robots and Systems, {IROS} 2018, Madrid, Spain, October 1-5, 2018}, pages
  1--8. {IEEE}, 2018.

\bibitem{DBLP:journals/corr/abs-1907-10326}
Jin~Han Lee, Myung{-}Kyu Han, Dong~Wook Ko, and Il~Hong Suh.
\newblock From big to small: Multi-scale local planar guidance for monocular
  depth estimation.
\newblock {\em CoRR}, abs/1907.10326, 2019.

\bibitem{DBLP:journals/corr/abs-2012-12397}
Ming Liang, Bin Yang, Yun Chen, Rui Hu, and Raquel Urtasun.
\newblock Multi-task multi-sensor fusion for 3d object detection.
\newblock {\em CoRR}, abs/2012.12397, 2020.

\bibitem{DBLP:journals/pami/LinGGHD20}
Tsung{-}Yi Lin, Priya Goyal, Ross~B. Girshick, Kaiming He, and Piotr
  Doll{\'{a}}r.
\newblock Focal loss for dense object detection.
\newblock {\em {IEEE} Trans. Pattern Anal. Mach. Intell.}, 42(2):318--327,
  2020.

\bibitem{DBLP:journals/corr/abs-2112-11088}
Zhe Liu, Tengteng Huang, Bingling Li, Xiwu Chen, Xi Wang, and Xiang Bai.
\newblock Epnet++: Cascade bi-directional fusion for multi-modal 3d object
  detection.
\newblock {\em CoRR}, abs/2112.11088, 2021.

\bibitem{DBLP:conf/cvpr/NohLH21}
Jongyoun Noh, Sanghoon Lee, and Bumsub Ham.
\newblock {HVPR:} hybrid voxel-point representation for single-stage 3d object
  detection.
\newblock In {\em {IEEE} Conference on Computer Vision and Pattern Recognition,
  {CVPR} 2021, virtual, June 19-25, 2021}, pages 14605--14614. Computer Vision
  Foundation / {IEEE}, 2021.

\bibitem{DBLP:conf/cvpr/PanXSLH21}
Xuran Pan, Zhuofan Xia, Shiji Song, Li~Erran Li, and Gao Huang.
\newblock 3d object detection with pointformer.
\newblock In {\em {IEEE} Conference on Computer Vision and Pattern Recognition,
  {CVPR} 2021, virtual, June 19-25, 2021}, pages 7463--7472. Computer Vision
  Foundation / {IEEE}, 2021.

\bibitem{2017Automatic}
A. Paszke, S. Gross, S. Chintala, G. Chanan, E. Yang, Z. Devito, Z. Lin, A.
  Desmaison, L. Antiga, and A. Lerer.
\newblock Automatic differentiation in pytorch.
\newblock 2017.

\bibitem{DBLP:conf/cvpr/QiCLG20}
Charles~R. Qi, Xinlei Chen, Or Litany, and Leonidas~J. Guibas.
\newblock Imvotenet: Boosting 3d object detection in point clouds with image
  votes.
\newblock In {\em 2020 {IEEE/CVF} Conference on Computer Vision and Pattern
  Recognition, {CVPR} 2020, Seattle, WA, USA, June 13-19, 2020}, pages
  4403--4412. Computer Vision Foundation / {IEEE}, 2020.

\bibitem{DBLP:conf/cvpr/QiLWSG18}
Charles~R. Qi, Wei Liu, Chenxia Wu, Hao Su, and Leonidas~J. Guibas.
\newblock Frustum pointnets for 3d object detection from {RGB-D} data.
\newblock In {\em 2018 {IEEE} Conference on Computer Vision and Pattern
  Recognition, {CVPR} 2018, Salt Lake City, UT, USA, June 18-22, 2018}, pages
  918--927. Computer Vision Foundation / {IEEE} Computer Society, 2018.

\bibitem{DBLP:conf/nips/QiYSG17}
Charles~Ruizhongtai Qi, Li Yi, Hao Su, and Leonidas~J. Guibas.
\newblock Pointnet++: Deep hierarchical feature learning on point sets in a
  metric space.
\newblock In Isabelle Guyon, Ulrike von Luxburg, Samy Bengio, Hanna~M. Wallach,
  Rob Fergus, S.~V.~N. Vishwanathan, and Roman Garnett, editors, {\em Advances
  in Neural Information Processing Systems 30: Annual Conference on Neural
  Information Processing Systems 2017, December 4-9, 2017, Long Beach, CA,
  {USA}}, pages 5099--5108, 2017.

\bibitem{DBLP:journals/corr/abs-2004-03080}
Rui Qian, Divyansh Garg, Yan Wang, Yurong You, Serge~J. Belongie, Bharath
  Hariharan, Mark Campbell, Kilian~Q. Weinberger, and Wei{-}Lun Chao.
\newblock End-to-end pseudo-lidar for image-based 3d object detection.
\newblock {\em CoRR}, abs/2004.03080, 2020.

\bibitem{DBLP:conf/cvpr/ReadingHCW21}
Cody Reading, Ali Harakeh, Julia Chae, and Steven~L. Waslander.
\newblock Categorical depth distribution network for monocular 3d object
  detection.
\newblock In {\em {IEEE} Conference on Computer Vision and Pattern Recognition,
  {CVPR} 2021, virtual, June 19-25, 2021}, pages 8555--8564. Computer Vision
  Foundation / {IEEE}, 2021.

\bibitem{DBLP:conf/cvpr/ShiWL19}
Shaoshuai Shi, Xiaogang Wang, and Hongsheng Li.
\newblock Pointrcnn: 3d object proposal generation and detection from point
  cloud.
\newblock In {\em {IEEE} Conference on Computer Vision and Pattern Recognition,
  {CVPR} 2019, Long Beach, CA, USA, June 16-20, 2019}, pages 770--779. Computer
  Vision Foundation / {IEEE}, 2019.

\bibitem{DBLP:journals/pami/ShiWSWL21}
Shaoshuai Shi, Zhe Wang, Jianping Shi, Xiaogang Wang, and Hongsheng Li.
\newblock From points to parts: 3d object detection from point cloud with
  part-aware and part-aggregation network.
\newblock {\em {IEEE} Trans. Pattern Anal. Mach. Intell.}, 43(8):2647--2664,
  2021.

\bibitem{DBLP:conf/icra/SindagiZT19}
Vishwanath~A. Sindagi, Yin Zhou, and Oncel Tuzel.
\newblock Mvx-net: Multimodal voxelnet for 3d object detection.
\newblock In {\em International Conference on Robotics and Automation, {ICRA}
  2019, Montreal, QC, Canada, May 20-24, 2019}, pages 7276--7282. {IEEE}, 2019.

\bibitem{DBLP:journals/corr/abs-2005-13423}
Yunlei Tang, Sebastian Dorn, and Chiragkumar Savani.
\newblock Center3d: Center-based monocular 3d object detection with joint depth
  understanding.
\newblock {\em CoRR}, abs/2005.13423, 2020.

\bibitem{openpcdet2020}
OpenPCDet~Development Team.
\newblock Openpcdet: An open-source toolbox for 3d object detection from point
  clouds.
\newblock \url{https://github.com/open-mmlab/OpenPCDet}, 2020.

\bibitem{DBLP:journals/corr/VaswaniSPUJGKP17}
Ashish Vaswani, Noam Shazeer, Niki Parmar, Jakob Uszkoreit, Llion Jones,
  Aidan~N. Gomez, Lukasz Kaiser, and Illia Polosukhin.
\newblock Attention is all you need.
\newblock {\em CoRR}, abs/1706.03762, 2017.

\bibitem{DBLP:conf/cvpr/VoraLHB20}
Sourabh Vora, Alex~H. Lang, Bassam Helou, and Oscar Beijbom.
\newblock Pointpainting: Sequential fusion for 3d object detection.
\newblock In {\em 2020 {IEEE/CVF} Conference on Computer Vision and Pattern
  Recognition, {CVPR} 2020, Seattle, WA, USA, June 13-19, 2020}, pages
  4603--4611. Computer Vision Foundation / {IEEE}, 2020.

\bibitem{DBLP:conf/cvpr/Wang0ZY21}
Chunwei Wang, Chao Ma, Ming Zhu, and Xiaokang Yang.
\newblock Pointaugmenting: Cross-modal augmentation for 3d object detection.
\newblock In {\em {IEEE} Conference on Computer Vision and Pattern Recognition,
  {CVPR} 2021, virtual, June 19-25, 2021}, pages 11794--11803. Computer Vision
  Foundation / {IEEE}, 2021.

\bibitem{DBLP:conf/icra/WangWLTCS19}
Tsun{-}Hsuan Wang, Fu{-}En Wang, Juan{-}Ting Lin, Yi{-}Hsuan Tsai, Wei{-}Chen
  Chiu, and Min Sun.
\newblock Plug-and-play: Improve depth prediction via sparse data propagation.
\newblock In {\em International Conference on Robotics and Automation, {ICRA}
  2019, Montreal, QC, Canada, May 20-24, 2019}, pages 5880--5886. {IEEE}, 2019.

\bibitem{DBLP:conf/cvpr/WangTF20}
Weiyao Wang, Du Tran, and Matt Feiszli.
\newblock What makes training multi-modal classification networks hard?
\newblock In {\em 2020 {IEEE/CVF} Conference on Computer Vision and Pattern
  Recognition, {CVPR} 2020, Seattle, WA, USA, June 13-19, 2020}, pages
  12692--12702. Computer Vision Foundation / {IEEE}, 2020.

\bibitem{DBLP:conf/cvpr/WangCGHCW19}
Yan Wang, Wei{-}Lun Chao, Divyansh Garg, Bharath Hariharan, Mark~E. Campbell,
  and Kilian~Q. Weinberger.
\newblock Pseudo-lidar from visual depth estimation: Bridging the gap in 3d
  object detection for autonomous driving.
\newblock In {\em {IEEE} Conference on Computer Vision and Pattern Recognition,
  {CVPR} 2019, Long Beach, CA, USA, June 16-20, 2019}, pages 8445--8453.
  Computer Vision Foundation / {IEEE}, 2019.

\bibitem{DBLP:journals/corr/abs-1903-01864}
Zhixin Wang and Kui Jia.
\newblock Frustum convnet: Sliding frustums to aggregate local point-wise
  features for amodal 3d object detection.
\newblock {\em CoRR}, abs/1903.01864, 2019.

\bibitem{DBLP:conf/iccvw/WengK19}
Xinshuo Weng and Kris Kitani.
\newblock Monocular 3d object detection with pseudo-lidar point cloud.
\newblock In {\em 2019 {IEEE/CVF} International Conference on Computer Vision
  Workshops, {ICCV} Workshops 2019, Seoul, Korea (South), October 27-28, 2019},
  pages 857--866. {IEEE}, 2019.

\bibitem{wu2019detectron2}
Yuxin Wu, Alexander Kirillov, Francisco Massa, Wan-Yen Lo, and Ross Girshick.
\newblock Detectron2.
\newblock \url{https://github.com/facebookresearch/detectron2}, 2019.

\bibitem{DBLP:journals/corr/abs-1911-06084}
Liang Xie, Chao Xiang, Zhengxu Yu, Guodong Xu, Zheng Yang, Deng Cai, and
  Xiaofei He.
\newblock {PI-RCNN:} an efficient multi-sensor 3d object detector with
  point-based attentive cont-conv fusion module.
\newblock {\em CoRR}, abs/1911.06084, 2019.

\bibitem{DBLP:conf/cvpr/XuAJ18}
Danfei Xu, Dragomir Anguelov, and Ashesh Jain.
\newblock Pointfusion: Deep sensor fusion for 3d bounding box estimation.
\newblock In {\em 2018 {IEEE} Conference on Computer Vision and Pattern
  Recognition, {CVPR} 2018, Salt Lake City, UT, USA, June 18-22, 2018}, pages
  244--253. Computer Vision Foundation / {IEEE} Computer Society, 2018.

\bibitem{DBLP:journals/sensors/YanML18}
Yan Yan, Yuxing Mao, and Bo Li.
\newblock {SECOND:} sparsely embedded convolutional detection.
\newblock {\em Sensors}, 18(10):3337, 2018.

\bibitem{DBLP:journals/corr/abs-1906-06310}
Yurong You, Yan Wang, Wei{-}Lun Chao, Divyansh Garg, Geoff Pleiss, Bharath
  Hariharan, Mark Campbell, and Kilian~Q. Weinberger.
\newblock Pseudo-lidar++: Accurate depth for 3d object detection in autonomous
  driving.
\newblock {\em CoRR}, abs/1906.06310, 2019.

\bibitem{DBLP:journals/corr/abs-2204-00325}
Yanan Zhang, Jiaxin Chen, and Di Huang.
\newblock Cat-det: Contrastively augmented transformer for multi-modal 3d
  object detection.
\newblock {\em CoRR}, abs/2204.00325, 2022.

\bibitem{DBLP:journals/corr/abs-2012-09164}
Hengshuang Zhao, Li Jiang, Jiaya Jia, Philip H.~S. Torr, and Vladlen Koltun.
\newblock Point transformer.
\newblock {\em CoRR}, abs/2012.09164, 2020.

\bibitem{DBLP:conf/aaai/ZhaoLHH19}
Xin Zhao, Zhe Liu, Ruolan Hu, and Kaiqi Huang.
\newblock 3d object detection using scale invariant and feature reweighting
  networks.
\newblock In {\em The Thirty-Third {AAAI} Conference on Artificial
  Intelligence, {AAAI} 2019, The Thirty-First Innovative Applications of
  Artificial Intelligence Conference, {IAAI} 2019, The Ninth {AAAI} Symposium
  on Educational Advances in Artificial Intelligence, {EAAI} 2019, Honolulu,
  Hawaii, USA, January 27 - February 1, 2019}, pages 9267--9274. {AAAI} Press,
  2019.

\bibitem{DBLP:conf/cvpr/ZhengTJF21}
Wu Zheng, Weiliang Tang, Li Jiang, and Chi{-}Wing Fu.
\newblock {SE-SSD:} self-ensembling single-stage object detector from point
  cloud.
\newblock In {\em {IEEE} Conference on Computer Vision and Pattern Recognition,
  {CVPR} 2021, virtual, June 19-25, 2021}, pages 14494--14503. Computer Vision
  Foundation / {IEEE}, 2021.

\bibitem{DBLP:journals/corr/abs-1711-06396}
Yin Zhou and Oncel Tuzel.
\newblock Voxelnet: End-to-end learning for point cloud based 3d object
  detection.
\newblock {\em CoRR}, abs/1711.06396, 2017.

\end{thebibliography}
}
	
\end{document}